\documentclass[11pt]{article}
\usepackage{tabu}
\usepackage{array}
\usepackage{amsmath}
\usepackage{amsthm}
\usepackage{amssymb}
\usepackage{algorithm}
\usepackage{subfig}
\usepackage{dirtytalk}
\usepackage{color}
\usepackage[english]{babel}
\usepackage{graphicx}
\usepackage{grffile}
\usepackage{wrapfig,epsfig}
\usepackage{epstopdf}
\usepackage{url}
\usepackage{color}
\usepackage{epstopdf}
\usepackage{algpseudocode}
\usepackage[T1]{fontenc}
\usepackage{bbm}
\usepackage{comment}
\usepackage{dsfont}
\usepackage{mathtools}
\usepackage{enumitem}
\usepackage{mathtools}
\usepackage{amsmath}
\usepackage{multirow}
\usepackage{booktabs}
\usepackage{wrapfig}
\usepackage{multicol}
\usepackage{comment} 
\usepackage{tabularx,booktabs}
\usepackage{wrapfig,lipsum,booktabs}
\usepackage{xcolor}
\usepackage{multirow}
\usepackage{graphicx}

\usepackage{tikz}
\usepackage{hyperref}  
\hypersetup{
    colorlinks=true,
    citecolor=red,
    linkcolor=blue,
    filecolor=cyan,
    urlcolor=magenta,
}
\usetikzlibrary{arrows}
\usepackage[margin=1in]{geometry}
\graphicspath{{./figs/}}

\usepackage[nameinlink,capitalize]{cleveref}
\usepackage{textcomp}
\usepackage{multirow}
\newtheorem{theorem}{Theorem}[section]

\newtheorem{definition}[theorem]{Definition}

\newcommand{\wh}{\widehat}

\renewcommand{\hat}{\wh}

\definecolor{mygreen}{RGB}{80,180,0}
\definecolor{b2}{RGB}{51,153,255}

\newcommand{\nc}{\newcommand}

\nc{\Ra}{\Rightarrow}
\nc{\zo}{\{0,1\}}

\title{Estimating and Improving Fairness with Adversarial Learning}

\author{
Xiaoxiao Li\thanks{\texttt{xl32@princeton.edu}. Princeton University.} \hspace{5mm}
\and
Ziteng Cui\thanks{\texttt{cuiziteng@sjtu.edu.cn }. Shanghai Jiaotong University.} \hspace{5mm}
\and
Yifan Wu\thanks{\texttt{yfwu@seas.upenn.edu }. University of Pennsylvania.} \hspace{5mm}
\and
Lin Gu\thanks{\texttt{lin.gu@riken.jp}. RIKEN AIP, University of Tokyo.} \hspace{5mm}
\and
Tatsuya Harada \thanks{\texttt{harada@mi.t.u-tokyo.ac.jp}. University of Tokyo, RIKEN AIP} 
}
\date{}

\begin{document}
\begin{titlepage}
  \maketitle
  \begin{abstract}
Fairness and accountability are two essential pillars for trustworthy Artificial Intelligence (AI) in healthcare. However, the existing AI model may be biased in its decision marking. To tackle this issue, we propose an adversarial multi-task training strategy to simultaneously \textbf{mitigate} and \textbf{detect} bias in the deep learning-based medical image analysis system. Specifically, we propose to add a discrimination module against bias and a critical module that predicts unfairness within the base classification model. We further impose an orthogonality regularization to force the two modules to be independent during training. Hence, we can keep  these deep learning tasks distinct from one another, and avoid collapsing them into a singular point on the manifold. Through this adversarial training method, the data from the under-privileged group, which is vulnerable to bias because of attributes such as sex and skin tone, are transferred into a domain that is neutral relative to these attributes. Furthermore, the critical module can predict fairness scores for the data with unknown sensitive attributes. We evaluate our framework on a large-scale public-available skin lesion dataset under various fairness evaluation metrics. The experiments demonstrate the effectiveness of our proposed method for estimating and improving fairness in the deep learning-based medical image analysis system.
  \end{abstract}
  \thispagestyle{empty}
\end{titlepage}

\newpage

\section{Introduction}

The success of deep learning can be partially attributed to data-driven methodologies that automatically recognize patterns in large amounts of data. However, this end-to-end paradigm leaves the deep learning model vulnerable to biases in the model itself and to biases in the data: such as, user groups with \textit{sensitive (protected) attributes} (\textit{i.e.} age or sex) that are over or under represented in the deep learning model ~\cite{hardt2016equality,Du2020survey}. As a result, the deep learning models tend to replicate and even amplify this data bias. For example, a recent study revealed statistically significant differences in performance on medical imaging-based diagnoses using gender imbalanced datasets~\cite{larrazabal2020gender}. Existing chest X-ray classifiers were found exhibiting disparities in performance across subgroups distinguished by different phenotypes ~\cite{seyyedkalantari2020chexclusion,seyyedkalantari2020chexclusion}.  Similar to forms of discrimination linked to face recognition~\cite{wu2019privacy}, darker skinned patients may be under-presented~\cite{Kinyanjui20MICCAI} in existing dermatology datasets: ISIC 2018 Challenge dataset~\cite{tschandl2018ham10000,codella2018skin}. A further study~\cite{Abbasi2020SkinBias} demonstrated that the skin lesion algorithm could show variable performance relative to other under-represented subgroups. If the decision-making is (partially) based on the values of sensitive attributes, the consequences can be irreversible or even fatal, especially in medical applications.

Unfairness mitigations mainly fall in the following three areas: 1) adversarial learning~\cite{beutel2017data,wang2019balanced,zhang2018mitigating,xu2018fairgan}; 2) calibration~\cite{zhao2017men,hardt2016equality}; 3) incorporating priors into feature attribution~\cite{hendricks2018women,liu2019incorporating}. The adversarial learning approach is considered the most generalizable approach to different bias inducements. Although there are bias mitigation methods proposed in text and natural image analysis, how to mitigate the biases in medical imaging has not been well explored. Additionally, although there are over 70 fairness metrics~\cite{aif360-oct-2018}, they all require the explicit marking of sensitive protected attributes and the explicit labelling of classification tasks, neither of which may be explicitly identified in existing data sets in real healthcare applications.

To enhance the accountability and fairness of artificial intelligence, we propose an adversarial training-based framework. Unlike the existing bias mitigation framework, our proposed pipeline is the first to simultaneously \textbf{mitigate} and  \textbf{detect} biases in medical image analysis. As illustrated in Fig.\ref{fig:structure}, the \textit{discriminator} is comprised of two tasks: 1. the \textbf{bias discrimination module} distinguishes between privileged and unprivileged samples, and 2. the \textbf{fairness critical module} predicts the fairness scores of given data.  Through adversarial training with the discriminator block, the \textbf{feature generator} can generate discrimination-free features that serve as the inputs for  the \textbf{feature classifier}. 

\textbf{Our contributions} are summarized in four folds: 1) This work addresses the neglected bias problem in medical image analysis and provides bias mitigation solutions via the adversarial training of a discrimination module; 2)  To ensure the accountability of trustworthy AI, we propose to co-train a critical module that is able to estimate bias for the inference dataset without knowing the its labels and sensitive attributes; 3) To force the critical module to make fairness estimations independently of bias discrimination, we introduce a novel orthogonality regulation; 4) As an early effort towards fair AI design for medical image analysis, our framework is demonstrated to be effective in predicting and mitigating bias in medical imaging tasks. 
\section{Problem Setup}
\subsection{Problem Statement}
Given a pair of data $\{x,y\}$, where $x\in \mathbb{R}^d$ is the input image and $y\in \mathcal{C}$, where $\mathcal{C}=\{1,2,\dots,k\}$ and $k$ is the number of classes representing diagnose label, our framework aims to map input $x$ into an intermediate space $h=G(x)  \in \mathbb{R}^{d^\prime}$. The base classification algorithm $C(\cdot)$ then map the $h$ to the final prediction  $\hat{y} = C(G(x))$. Using threshold, we can map output $\hat{y}$ to class $c$, namely $f(x) \rightarrow c$, for $f=C(G(\cdot))$ and $c \in \mathcal{C}$. The intermediate representation $G(\cdot)$ should ensure the prediction $\hat{y}$ to suffer least biases with respect to the protected attribute $z$. We denote the data set $X$, sensitive feature set $Z$ and label set $Y$. In our setting, $z \in \mathcal{Z}=\{0,1\}$ is a binary sensitive feature that may induce bias. $z=0$ indicates privileged samples, vice versa,  $z=1$ indicates under-privileged samples. 

\subsection{Notation of Fairness}
\label{sec:notation}
 We treat the samples with sensitive features that benefit classification as privileged samples. On the contrary, the sample with sensitive features against classification are viewed as under-privileged samples.

\begin{definition}[Fair classifier] We define a classifier $f$ with respect to data distribution $Pr$ on $\{(X,Z,Y): \mathbb{R}^d \times \mathcal{Z} \times \mathcal{C}\}$, where $\mathcal{C}=\{1,2,\dots,k\}$ and $k$ is the number of classes. The classifier $f$ is fair if 
\begin{equation*}
    Pr(f(x,z)\in \mathcal{C}|z=0) = Pr(f(x,z)\in \mathcal{C}|z=1)
\end{equation*}
\end{definition}
In other words, if the distance between the two distribution of outputs given the two conditions, $z=0$ or $z= 1$, of the function is zero, then $f$ is denoted as fair classifier.  
We can have the shortened notation for fairness as \cite{chzhen2020fair}:
\begin{align}\label{eq:obj1}
    \mathcal{U}(f,\mathcal{C}) = |Pr(f(x,z)\in \mathcal{C}|z=0) - Pr(f(x,z)\in \mathcal{C}|z=1)|.
\end{align}
Instead of directly optimizing Eq.~\ref{eq:obj1}, we achieve the fairness while maintaining data utility through adversarial training (illustrated in Section~\ref{sec:method}). 

\subsubsection{Fairness measurements} 
\label{sec:fairmeasure}
 Given $x_i, y_i, z_i$ to be the input, label, bias indicator for the $i$th sample in dataset $\mathcal{D} = \{ X,Y,Z\}$. Let denote $\hat{y}_i$ as the predicted label of sample $i$. $Pr$ is the classification probability. The true positive and false positive rates are:\\
\begin{align*}
    TPR_{z} = \frac{|\{i|\hat{y}_i=y_i,z_i=z\}|}{|\{i|\hat{y}_i=y_i\}|} = Pr_{(x_i,y_i,z_i)\in \mathcal{D}}(\hat{y}_i=y_i|z_i=z) \\
    FPR_{z} = \frac{|\{i|\hat{y}_i \neq y_i,z_i=z\}|}{|\{i|\hat{y}_i\neq y_i\}|} = Pr_{(x_i,y_i,z_i)\in \mathcal{D}}(\hat{y}_i\neq y_i|z_i=z).
\end{align*}

 Following \cite{bellamy2018ai,dwork2012fairness,hardt2016equality}, we quantify the discrimination or the bias of the model using the fairness metrics \textit{Statistical Parity Difference (SPD), Equal opportunity difference (EOD), and Average Odds Difference (AOD)}:
\begin{itemize}
    \item \textit{Statistical Parity Difference (SPD).} SPD measures the difference in the probability of positive outcome between the privileged and under-privileged groups:
\begin{align*}
    SPD = Pr_{(x_i,y_i,z_i)\in \mathcal{D}}(\hat{y}_i=y_i|z_i=0) - Pr_{(x_i,y_i,z_i)\in \mathcal{D}}(\hat{y}_i=y_i|z_i=1).
\end{align*}
\item \textit{Equal opportunity difference (EOD).} EOD measures the difference in TPR for the privileged and under-privileged groups:
\begin{align*}
    EOD = TPR_{z=0}-TPR_{z=1}.
\end{align*}
\item \textit{- Average Odds Difference (AOD).} AOD is defined as the average of the difference in the false positive rates and true positive rates for under-privileged and privileged groups:
\begin{align*}
    AOD = \frac{1}{2} \left[ (FPR_{z=0} - FPR_{z=1}) + (TPR_{z=0}-TPR_{z=1})\right].
\end{align*}
\end{itemize}

\section{Methods}
\label{sec:method}
\subsection{Overview}
\begin{figure*}[t]
    \centering
    \includegraphics[width=0.99\textwidth]{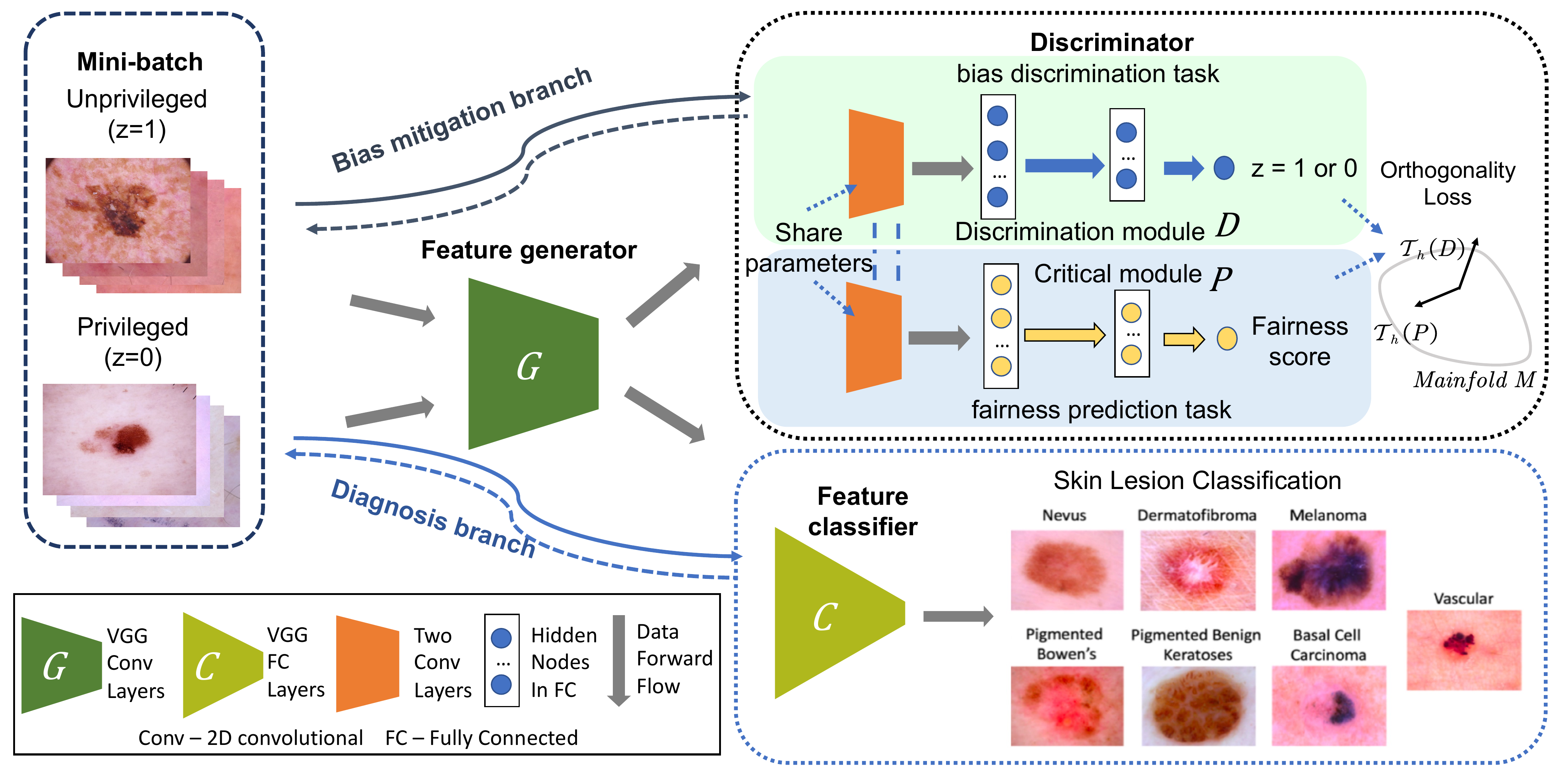}
    \caption{\small Overview of our proposed bias mitigation and fairness prediction pipeline.}
    \label{fig:structure}
\end{figure*}

Our pipeline is depicted in Fig. \ref{fig:structure}. The network architecture is composed of four modules. The feature generator $G$ and feature classifier $C$ are drawn from the vanilla classification model, forming the diagnosis branch. To mitigate bias and predict fairness for inference data, we introduce two modules, the bias discrimination module $D$ for bias detection, and critical module $P$ for fairness prediction. These two modules are on the side of the discriminator and adversarially trained with $G$ , forming the bias mitigation branch. To reduce the total number of parameters, $D$ and $P$ share the first few layers, then split into two branches with the same network architectures but with different parameters. 

\subsection{Mitigate Bias via Adversarial Training}

 Suppose the training samples can be divided in to privileged data $X^p$ (whose $z=0)$ and under-privileged data $X^u$ (whose $z=1$). For fair representation learning, we first initialize an adversarial training pipeline that we train a local feature extractor $G$ that takes images as inputs. $G$'s outputs are fed into disease classifier $C$ to predict $y$ and bias discrimination branch $D$ to predict $z$.

 Model optimization is divided into three independent steps. First, $G$ and $C$ are trained for the target task using cross entropy $\mathcal{L}_{ce}$. 
 Second, $G$ and $C$ are fixed, $D$ is trained to identify the sensitive attribute associated with the input (such as whether the input a male or femal). The objective function $L_{advD}$ for discriminator module $D$ is defined as:
\begin{equation} \label{eq:gan1}
     \mathcal{L}_{advD}(X^p,X^u,D)\! =\! - \mathbb{E}_{x^p \sim X^p}[\log D(G(x^p))]  - \mathbb{E}_{x^u \sim X^u}[\log (1-D(G(x^u)))].
\end{equation}
Then, $D$ is fixed. $G$ is further updated to confuse $D$ using $L_{advG}$ by encouraging the outputs of $D$ for arbitrary input to be 0 --- no bias in the feature space:
\begin{equation} \label{eq:gan2}
\mathcal{L}_{advG}(X^p, \! X^u,\! G) \! = \! - \mathbb{E}_{x^p \sim X^p}[\log (1\!-\!D(G(x^p)))] \!  - \! \mathbb{E}_{x^u \sim X^u}[\log (1\!-\!D(G(x^u)))].
\end{equation}

\begin{algorithm}[t]
    		\caption{Adversarial Training Pipeline}\label{ag1}
    		\hspace*{\algorithmicindent} \textbf{Input:} {1. $(X^p,Y^p)$, privileged samples; 2. $(X^u, Y^u)$, under-privileged samples; 3. $G$, feature generators; $C$, disease classifier; 4. $D$, bias discrimination module; 5. $P$, critic prediction module to predict fairness scores; 6. $K$, number of optimization iterations; 7. $B$, number of mini-batch.
    			\begin{algorithmic}[1]
    				\State {Initialize parameters $\{\theta_G, \theta_C, \theta_D, \theta_P\}$}
    				\For{$k = 1$ to $K$}
    				\For{$b = 1$ to $B$}
    				\State{Sample mini-batch $\{(X_b,Y_b)\}$  from $\{(X^p,Y^p)\}$ and $\{(X^u,Y^u)\}$}
    				\State{\textbf{Train disease classifier:}}
    				\State{Update  $\theta_{C}^{(k)}, \theta_{G}^{(k)}$ loss $\mathcal{L}_{ce}$ to update $\theta_{C}^{(k)}$ and $\theta_{G}^{(k)}$} \Comment{Cross-entropy loss.}
    				\State{\textbf{Adversarial training:}}
    				\State{Update $\theta_{D}^{(k)}, \theta_{P}^{(k)}$ with $\mathcal{L}_{advD}+\mathcal{L}_{P}+\mathcal{L}_{orth}$} 
    				\text{\quad $\triangleright$ Based on Eq.~\eqref{eq:gan1}~\eqref{eq:gan3}~\eqref{eq:gan4}.}
    				\State{Update $\theta_{G}^{(k)}$ with $\mathcal{L}_{advG}$}
    				\Comment{Based on Eq.~\eqref{eq:gan2}.}
    				\EndFor
    				\EndFor
    			\end{algorithmic}
    			\hspace*{\algorithmicindent} \textbf{Return: trained modules} $\{C,G,P,D\}$
    		}
\end{algorithm}

\subsection{Co-training Discriminator on Conditional and Critic Modules}
Further more, we are motivated that fairness may not be guaranteed by using the adversarial strategy proposed above if the testing data distribution is not the same as the training data. To avoid providing biased decision, we want to know if the model is fair to the testing data during inference stage, when data's labels and sensitive attributes are not available. Hence, we propose to co-train a critical module $P$, which predict the fairness scores for the given trained model and the testing data.
As a valid assumption in training, each mini-batch contains samples from both privileged and under-privileged samples.\footnote{This can be achieve by stratified sampling and random shuffling.} The label of training $P$ can be any fairness measurements described in Section~\ref{sec:fairmeasure}. Given a training batch $X_b$, we calculate its fairness measurement $SPD_b$ using $X_b$'s true and predicted labels at the previous epoch as the prediction label for $P$. We choose SPD score as the label because SPD is less sensitive than the alternatives \cite{savani2020intra}. 
Thus, with the fixed feature generator $G$, the loss function for predictor $P$ is expressed as:
\begin{equation} \label{eq:gan3}
\mathcal{L}_{P}(X_b,P) = \frac{1}{2} \mathbb{E}_{x \sim X_b}\|P(G(x)) -  SPD_b \|^2.
\end{equation}
During training, as the feature generator $G$ gradually adjusts the fairness in the embedded space, thus the critical module $P$ can learn various levels of fairness.

\subsubsection{Orthogonality Regularization} 
Unlike the classic adversarial training, (\textit{i.e.}, DCGAN \cite{radford2015unsupervised}), we propose a model equipped with an independent predictor (critic module $P$) taking the feature $h=G(x)$ output form generator for fairness prediction as the auxiliary task. 

Two bottlenecks of multi-task adversarial training are the collapse of the hidden space and mutually curbed interactions of different tasks~\cite{suteu2019regularizing,he2020novel,qi2018global,bansal2018can}. 
Because the bias discrimination module $D$ and critical module $P$ are co-training using the same loss, the model behavior can be dominated by one module. The extreme scenario could be $D$ directly copies $P$ if they have the same architecture and vice versa. Motivated by~\cite{suteu2019regularizing,qi2018global,bansal2018can} that avoid the coupling effect of multi-task training by regularizing the gradients, we design our orthogonality regularization as follows. We can describe the tangent space $\mathcal{T}_h$ for the gradients on the hidden space $h=G(x)$.

To prevent the collapse of the tangent space (namely zero gradients for all the coordinates) and ensure the independence of training discriminator module $D$ and critical module $P$, we impose a regularity condition that the two tangent vectors $\mathcal{T}_h(D) = \frac{\partial{D}}{\partial{h}} \in \mathbb{R}^{1\times d^\prime}$ and $\mathcal{T}_h(P) = \frac{\partial{P}}{\partial{h}} \in \mathbb{R}^{1\times d^\prime} $ for the bias discrimination task and the fairness prediction task to be orthogonal. Namely we want to achieve $\mathcal{T}_h(D) \perp \mathcal{T}_h(P)$. 

We denote ${\rm \bf J} = [\mathcal{T}_{h}(D);\mathcal{T}_h(P)]$ and define the orthogonality loss as
\begin{equation} \label{eq:gan4}
    \mathcal{L}_{orth}(X,D,P) = \mathbb{E}_{x \sim X} \left \| {\rm \bf J}_x^{\top}{\rm \bf J}_x - {\rm \bf I}_2 \right \|.
\end{equation}
$\mathcal{L}_{orth}$ is simultaneously optimized with  $\mathcal{L}_{advD}$ (Eq. \eqref{eq:gan2}) and $\mathcal{L}_{advP}$ (Eq. \eqref{eq:gan3}). Putting each module together, the training scheme is described in Algorithm \ref{ag1}.

\section{Experiment}
We evaluate our framework on skin lesion detection against sensitive attribute: \textit{age}, \textit{sex} and \textit{skin tone}. 
\subsection{Datasets and Overall Setup}
Our data is extracted from the open skin lesion analysis dataset, Skin ISIC 2018~\cite{tschandl2018ham10000,codella2018skin}. The dataset consists of 10015 images: 327 actinic keratosis (AKIEC), 514 basal cell carcinoma (BCC), 115 dermatofibroma (DF), 1113 melanoma (MEL), 6705 nevus (NV), 1099 pigmented benign keratosis (BKL), 142 vascular lesions (VASC). Each example is a RGB color image with size $450 \times 600$. The privileged and under-privileged groups of ISIC dataset are separated by \textit{age} ( $\geq$60 and $<$60), \textit{sex} (male and female) and \textit{skin tone} (dark and light skin). \footnote{The processing details for skin color labeling are described in Appendix~\ref{app:ita}.} 

We randomly split the dataset into a training set ($80\%$) and a test set ($20\%$) to evaluate our algorithms. All the experiments are repeated five trials with different random seeds for splitting. Batch size is set to 64. We use Adam optimizer~\cite{kingma2014adam} and learning rate (lr) for classification training ($G+C$) is 1e-3; while lr for discriminator ($D+P$) is 1e-4.



\subsection{Biases in the Initial Model Condition}
\label{sec:base}
First, we present the bias existed in the \textbf{vanilla model} which only contains feature extractor $G$ and feature classifier $C$ module. We use VGG11\cite{simonyan2014very} to classify skin lesions among the 7 classes (AKIEC, BCC, DF, MEL, NV, BKL, and VASC). The convolutional layers are treated as $G$ and the fully-connected layers serve as $C$.  The results were listed in the `Vanilla' row of Table \ref{tab:classify}, where we used different weighted metrics (precision, recall, F1-score) to evaluate the testing classification performance on the subgroups with different sensitive attributes in the format of mean(std) for five trials. We noticed the performance gaps between the subgroups, which reveals the biases. We denoted the group with better classification performance (such as female, $\rm age<60$, light skin) as privileged group ($z=0$) and the opposite group (such as male, $\rm age\geq60$, dark skin) as under-privileged group with bias ($z=1$).

\begin{table}[!t]
\centering
\caption{\small  Model performance in mean(std). Higher scores indicate better performance.}
\label{tab:classify}
\resizebox{0.95\textwidth}{!}{%
\begin{tabular}{l|c|ccc|c|ccc|c|ccc}
\hline
 & \textbf{Sex} & Precision & Recall & F1-score & \textbf{Age} & Precision & Recall & F1-score  & \textbf{Skin} & Precision & Recall & F1-score  \\ \hline
\multirow{4}{*}{Vanilla} & Male & 0.898 & 0.869 & 0.885 & $\geq 60$ & 0.812 & 0.798 & 0.802  & Dark & 0.852 & 0.789 & 0.810  \\
& & (0.007) &  (0.005)  &  (0.005)  &  &  (0.010)  &  (0.009) &  (0.010)   &  &  (0.015)  &  (0.031)  &  (0.026)   \\
 & Female & 0.934 & 0.918 & 0.901 & $ < 60$ & 0.948 & 0.922 & 0.931 & Light& 0.928 & 0.896 & 0.908  \\ 
& & (0.018) &  (0.016)  &  (0.015)  &  &  (0.003)  &  (0.006) &  (0.004)   &  &  (0.011)  &  (0.019)  &  (0.015)   \\ \hline
\multirow{4}{*}{\begin{tabular}[c]{@{}l@{}}Ours\\ w/o  $\mathcal{L}_{orth}$ \end{tabular}}  & Male & 0.908 & 0.900 & 0.903 & $\geq 60$ & 0.836 & 0.827 & 0.829  & Dark & 0.852 & 0.862 & 0.836 \\
& & (0.017) &  (0.014)  &  (0.014)  &  &  (0.025)  &  (0.025) &  (0.026)   &  &  (0.022)  &  (0.024)  &  (0.028)   \\
 & Female & 0.931 & 0.889 & 0.903  & $ < 60$ & 0.953 & 0.941 & 0.946 &  Light & 0.933 & 0.908 & 0.919  \\ 
 & & (0.012) &  (0.010)  &  (0.009)  &  &  (0.008)  &  (0.009) &  (0.009)   &  &  (0.013)  &  (0.019)  &  (0.015)   \\\hline
\multirow{4}{*}{\begin{tabular}[c]{@{}l@{}}Ours\\ w $\mathcal{L}_{orth}$\end{tabular}}  & Male & 0.898 & 0.885 & 0.892 & $\geq 60$ & 0.824 & 0.835 & 0.827  & Dark & 0.845 & 0.866 & 0.839  \\
& & (0.010) &  (0.013)  &  (0.010)  &  &  (0.006)  &  (0.010) &  (0.009)   &  &  (0.004)  &  (0.007)  &  (0.011)   \\
 & Female & 0.890 & 0.900 & 0.895 & $ < 60$ & 0.924 & 0.929 & 0.912 & Light & 0.923 & 0.910 & 0.914  \\ 
 & & (0.016) &  (0.006)  &  (0.005)  &  &  (0.011)  &  (0.017) &  (0.006)   &  &  (0.005)  &  (0.010)  &  (0.004)   \\\hline
\end{tabular}%
}
\end{table}

\begin{table}[!t]
\centering
\caption{\small Fairness scores and correlations (Corr) of predictions in mean(std). Lower SPD, EOD and AOD indicate less bias. Hight Corr indicate better fairness prediction.}
\label{tab:pred}
\resizebox{0.9\textwidth}{!}{%
\begin{tabular}{l|cccc|cccc|cccc}
\toprule
 & \multicolumn{4}{c}{\textbf{Sex} ($\times 10^{-2}$ )} & \multicolumn{4}{|c}{\textbf{Age} ($\times 10^{-2}$ )}  & \multicolumn{4}{|c}{\textbf{Skin Tone} ($\times 10^{-2}$ )} \\ \cline{2-13} 
 & SPD & EOD & AOD & Corr & SPD & EOD & AOD & Corr & SPD & EOD & AOD & Corr \\ \hline
\multirow{2}{*}{Vanilla} & 8.3  & 8.8  & 10.2  & - & 12.8  & 9.6  & 7.8  & - & 10.2  & 11.3 & 8.6 & - \\ 
 & (2.9) & (1.6) & (2.9) & - & (3.2) & (3.0) & (1.7) & - &  (3.5) & (2.9) & (2.9) & - \\\hline
\multirow{2}{*}{Ours w/o  $\mathcal{L}_{orth}$} & 8.0  & 6.6  & \textbf{6.1}  & 13.1 & 11.2 & 9.8  & 9.3  & 24.4 & 7.6 & 13.9 & 12.4 & 15.3 \\ 
&(4.0) & (3.8) &(4.1) & (24.8) & (2.4) & (5.3) &  (4.2) & (12.2) & (4.8) & (2.2) & (5.6) & (12.4) \\\hline
\multirow{2}{*}{Ours w/ $\mathcal{L}_{orth}$} & \textbf{2.4} & \textbf{6.2} & 6.3 & \textbf{38.8}  & \textbf{6.9}  & \textbf{7.9}  & \textbf{5.7} & \textbf{32.6}  & \textbf{6.8}  & \textbf{6.9} & \textbf{7.3} & \textbf{44.3}  \\ 
& (0.6) & (2.1) & (1.8) & (6.0) & (0.4) &  (1.1) & (1.3) & (6.6) &  (0.8) &  (1.9) &  (1.4) & (11.3) \\\bottomrule
\end{tabular}%
}
\end{table}

\subsection{Bias Mitigation and Prediction}
\paragraph{Bias mitigation strategies.} Now we present our main experimental study by adding our bias mitigation strategies with the bias discrimination module $D$ and critical module $P$. As shown in Fig.~\ref{fig:structure}, the shared network of bias $D$ and $P$ is a two layer convolutional neural network, with channel numbers of 16 and 32, and kernel size of 3. The split branches of $D$ and $P$ are both two-layer FC networks (128-32-1), but will be updated independently. For the bias discrimination module $D$, we apply a sigmoid function to the outputs. For the critical module $P$, we average the output in the minibatch as the predicted fairness score. We followed the training strategy described in Algorithm~\ref{ag1}.
\paragraph{Fairness and classification performance.} We quantitatively measure the fairness scores (SPD, EOD, AOD) on the testing set for our models and the vanilla model, as shown in Table~\ref{tab:pred}. Lower fairness scores indicate the model is
less biased. All results are reported in the mean(std) format for five trials. Also, we conduct \textbf{ablation study} on orthogonality regularization $\mathcal{L}_{orth}$ (Eq~\ref{eq:gan4}). We name our adversarial learning scheme with and without orthogonality as `Ours w/ $\mathcal{L}_{orth}$' and `Ours w/o $\mathcal{L}_{orth}$' correspondingly.   Compared with the vanilla model, `Ours w/ $\mathcal{L}_{orth}$' achieves consistently lower fairness scores on all the above three fairness metrics, which suggests our model is less biased. Whereas, `Ours w/o $\mathcal{L}_{orth}$' perform worse in bias mitigation than `Ours w/ $\mathcal{L}_{orth}$.' Furthermore, as shown in Table~\ref{tab:classify}, compared with the vanilla model the vanilla model, our schemes preserve the utility of the data and even boost model classification performance.
Overall, `Ours w/o $\mathcal{L}_{orth}$' slightly but not significantly outperforms `Ours w/ $\mathcal{L}_{orth}$' on classification, which may imply a trade of adding orthogonality regularization.

\begin{figure*}[!t]
    \captionsetup[subfigure]{justification=centering}
    \centering
    \subfloat[\small Sex ]{\includegraphics[width=0.33\linewidth]{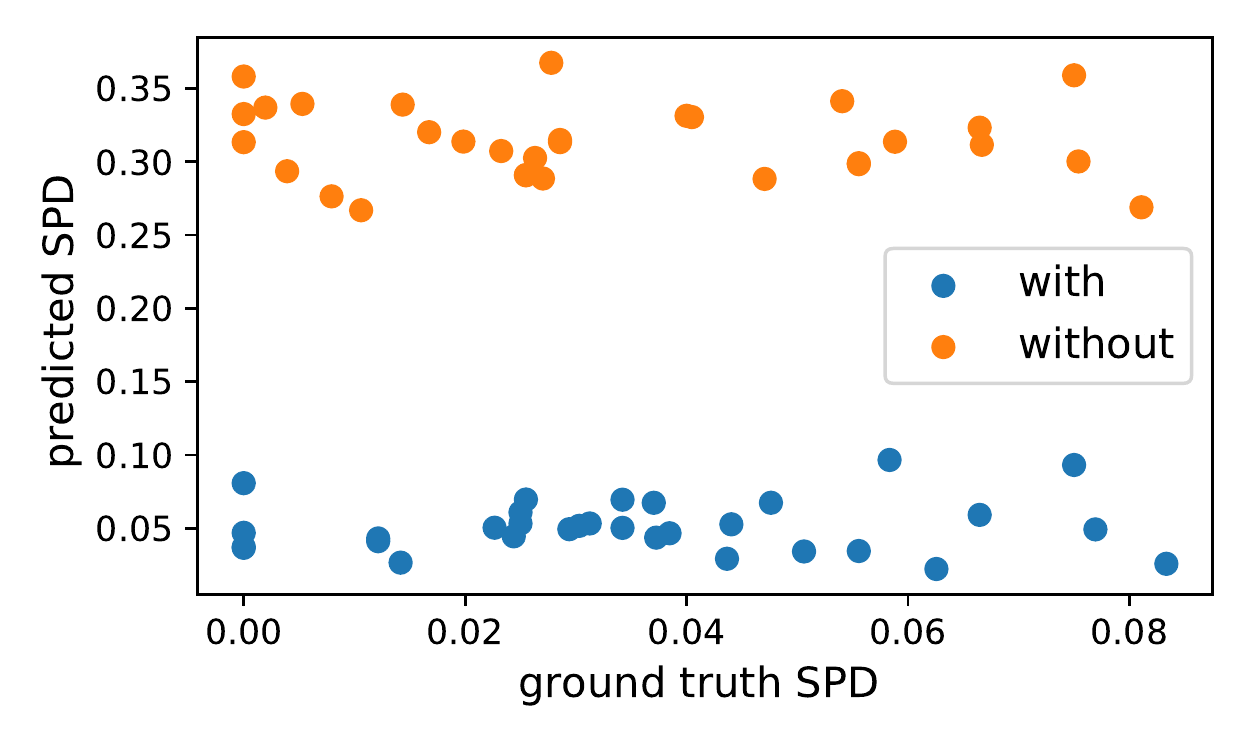}}
    \subfloat[\small Age ]{\includegraphics[width=0.33\linewidth]{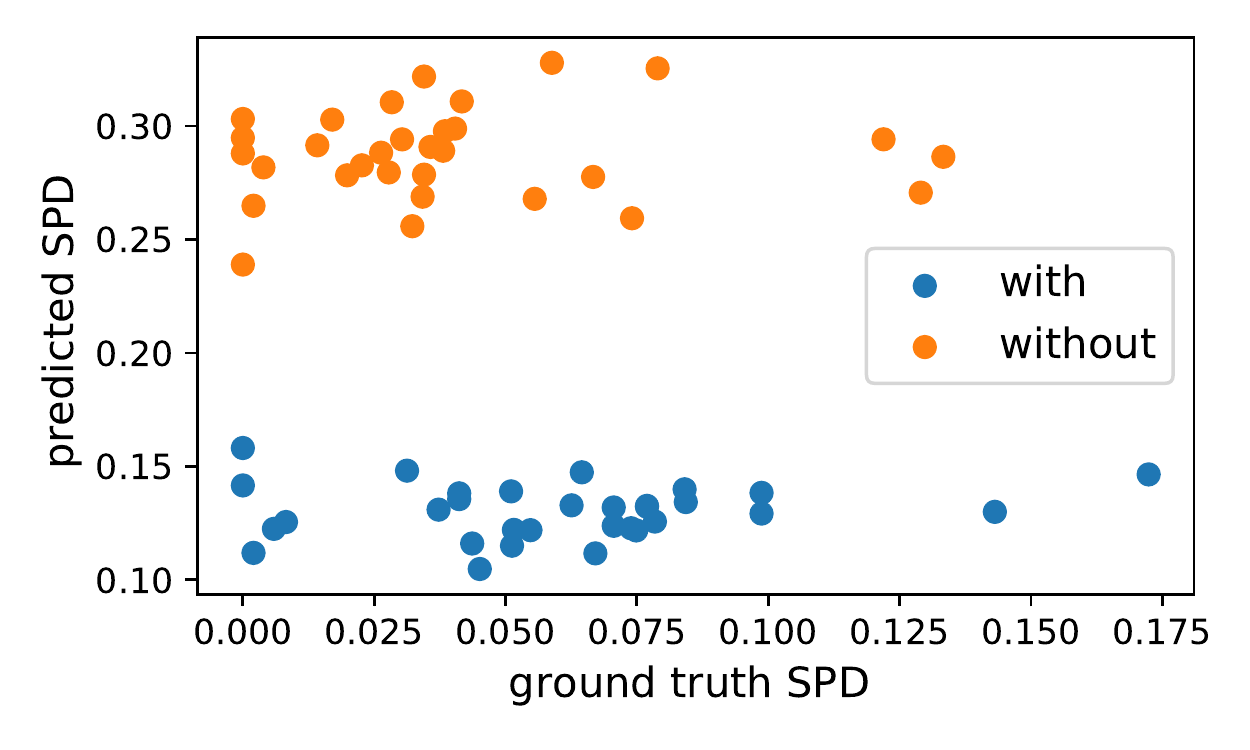}}
        \subfloat[\small Skin tone ]{\includegraphics[width=0.33\linewidth]{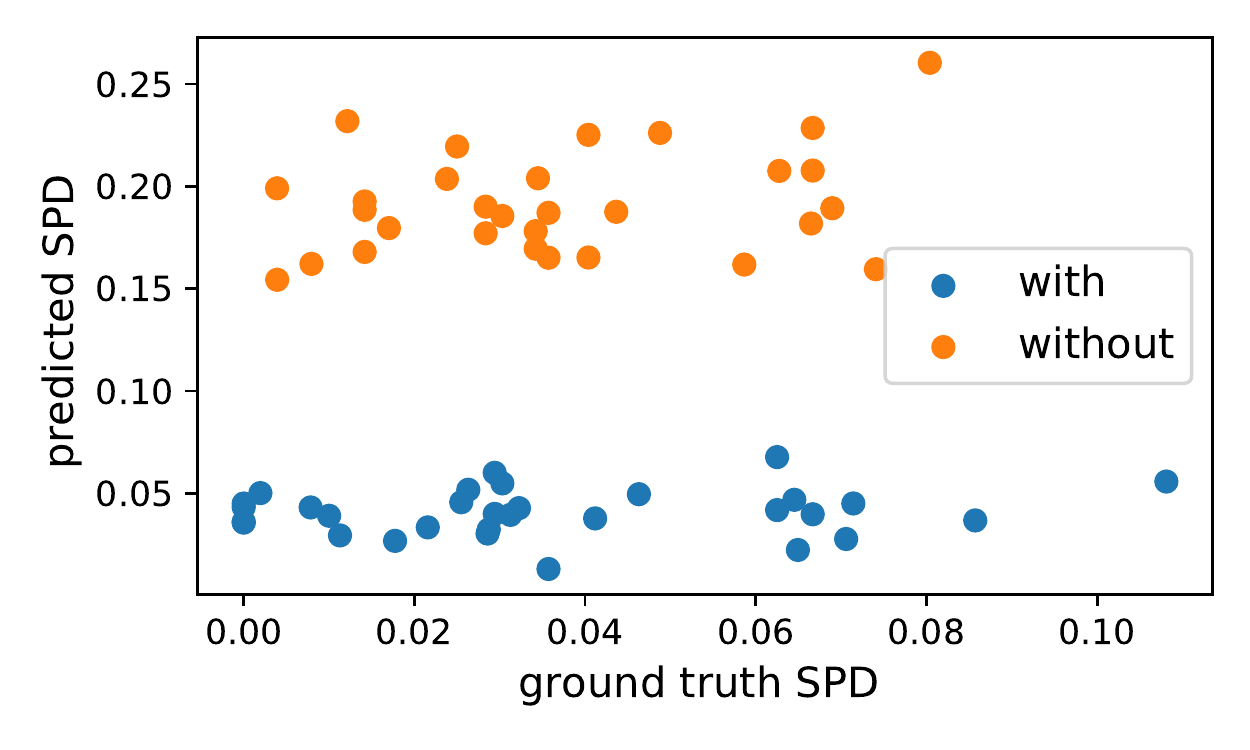}}
    \caption{\small Fairness score (SPD) prediction for the batches of testing dataset (randomly selecting the same splitting for all) with models debiased  on different sensitive attributes. Note that x-axis and y-axis have different scales and intervals. }
    \label{fig:viz}
\end{figure*}
\paragraph{Fairness prediction} Last, we show the effectiveness of the critical module $P$ for fairness prediction on the testing dataset batches. Each batch contains 64 images drawn from both privileged and under-privileged groups. The \textit{predicted SPD scores} are generated from $P$. We calculate the  \textit{ground truth SPD scores} based on classifier $C$ and ground truth classification labels for evaluation.\footnote{Note that our prediction $P(x)$ does not require knowing inference data's labels and sensitive attributes in practice if we do not evaluate prediction correctness.} The Pearson correlation between the ground truth SPD and predicted SPD are presented in the `Corr' column of Table~\ref{tab:pred}. Our method with $\mathcal{L}_{orth}$ achieves significant \textbf{more correlated} fairness prediction than without $\mathcal{L}_{orth}$. Furthermore, we show predicted and true SPD with respect to different sensitive attributes in Fig. \ref{fig:viz}. Our method with $\mathcal{L}_{orth}$ predict the \textbf{closer} fairness scores to ground truth than those without $\mathcal{L}_{orth}$. For example, in the task of predicting model fairness with respect to sex, predicted SPD scores with $\mathcal{L}_{orth}$ and ground truth \textbf{both} fall in the range [0, 0.1], while scores predicted without $\mathcal{L}_{orth}$ fall in range [0.25, 0.38].

\paragraph{Remark}
Compared to the vanilla model, our scheme can reduce biases in classifications and predict fairness scores for inference data, while keeping comparable classification performance. The orthogonality regularization improves bias mitigation and fairness prediction in multi-task adversarial training.

%
\section{Discussion and Conclusion}
In this work, we take the first step toward reducing bias in  deep learning algorithms with clinical applications. Specifically, we propose an adversarial training strategy that co-trains a bias discrimination module with the vanilla classifier. Although our adversarial learning strategy shows effectiveness in bias mitigation in comparison with various fairness measurements, we cannot conclude that the model is fair for arbitrary testing cohorts. In compliance with these limitations, we propose a novel fairness prediction scheme (critical module) to estimate the model fairness for a group of inference data. Therefore, model users can evaluate if the deployed model meets their fairness requirements. To this end, our work provides a promising avenue for future research seeking to thoroughly mitigate bias and evaluate the fairness of the model for incoming inference data.

\bibliographystyle{splncs04}
\bibliography{ref}

\newpage
\appendix
\section*{Appendix}
\subsection*{Roadmap} 
We list the notations table in Section~\ref{app:notation}.
The details of data preprocessing are in Section~\ref{app:ita}. 
\section{Notation Table}
\label{app:notation}
\vspace{-6mm}
\begin{table}[h]
\centering
\caption{\small Notations used in the paper.}
\label{tab:notation}
\resizebox{\linewidth}{!}{%
\begin{tabular}{c|l}
\hline
Notations & \multicolumn{1}{|c}{Descriptions} \\ \hline
$x,y,z$ & input image $x\in X$, classification label $y \in Y$, binary sensitive feature $z \in Z$\\
$X^p,X^u$ & privileged images (whose $z=0$), unprivileged images (whose $z=1$)\\
$X_b,Y_b$ & mini-batch images and labels\\
$\mathcal{D}$ & data set, where $\mathcal{D}=\{X,Y,Z\}$ \\
$\mathcal{Z}$ & sensitive feature space, where $z\in\mathcal{Z}=\{0,1\}$ \\
$k$ & the number of classes \\
$\mathcal{C}$ & classification label space, where $\mathcal{C}=\{1,\dots,k\}$\\
$G$ & feature generator that takes $x$ as input \\
$C$ & disease classifier that takes $G(x)$ as input \\
$D$ & bias discrimination module that takes $G({X}_b)$ as input and outputs bias detection result (0 or 1)\\
$P$ & critical predict module that takes $G({X}_b)$ as input and predicts batch-wise fairness score\\
$f$ & classifier $f(x) = \hat{y}$, with $f=C(G(\cdot))$ and $\hat{y} \in \mathcal{C}$ \\
$Pr$ & probability \\
$TPR$ & true positive rate \\
$FPR$ & false positive rate \\
$SPD$ & statistical parity difference \\
$SPD_b$ & statistical parity difference score calculated on batch $b$\\
$EOD$ & equal opportunity difference \\
$AOD$ & averaged odds difference \\
$h$ & hidden feature space, where $h=G(x)$ \\
$\mathcal{T}_h(D),\mathcal{T}_h(P)$ & gradient tangent planes \\
$\mathbf{J}$ & Jacobian matrix \\
$\mathbf{I}_N$ & $N$ by $N$ identity matrix\\
$\mathcal{L}_{ce}$ & cross-entropy loss \\
$\mathcal{L}_{advG}$ & adversarial loss for updating feature generator\\
$\mathcal{L}_{advD}$ & adversarial loss for updating discriminator block\\
$\mathcal{L}_{P}$ & prediction loss\\
$\mathcal{L}_{orth}$ & orthogonality regularization\\
\hline
\end{tabular}%
}
\vspace{-5mm}
\end{table}

\newpage
\section{Data Prepossessing}
\label{app:ita}
\begin{figure*}[h]
    \centering
    \includegraphics[width=10cm]{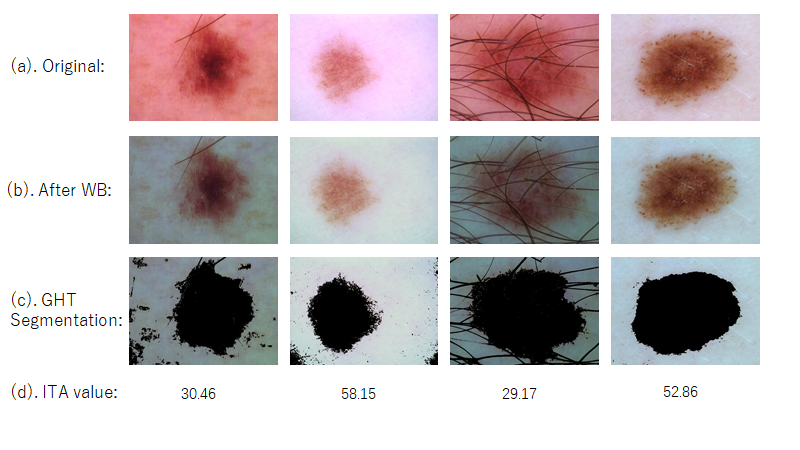}
    \caption{\small ITA value calculation:(a). the original image, (b). use white balance algorithm to reduce the influence of light, (c). image segmentation (d). calculate the ITA value}
    \label{ITA}
\end{figure*}

We rescale the images to $224 \times 224$ by center cropping and resizing. The privileged and unprivileged groups \textit{skin tone} is based on individual typology angle (ITA) criteria. We first use gray world white balance algorithm~\cite{huo2006robust} to correct the influence of light. Then we utilize the recent Generalized Histogram Thresholding (GHT)\cite{BarronECCV2020} method for segmentation, which kept surrounding skin of the lesions only. We followed \cite{Kinyanjui20MICCAI} transformed images from RGB-space to CIELab-space to calculate the ITA value of the non-lesion area. We set skins with ITA less than 45 as the dark skins. We set skins with ITA higher than 45 as the light skins. The description of this process is shown in Fig.\ref{ITA}. Based on age, 36.2\% of subjects are $\geq$ 60, and 63.8\% of subjects are $<$ 60. Based on sex, 54.3\% of subjects are male, and 45.7\% of subjects are female. Based on skin tone, 43.2\% of subjects are dark skin and 63.8\% of subjects are light skin.

\end{document}